%% file: main.tex
\documentclass[11pt,a4paper]{article}
\input{0-preamble}
\usepackage{xcolor}
\usepackage{booktabs}

\begin{document}
\maketitle
\begin{abstract}
We present FRMT, a new dataset and evaluation benchmark for Few-shot Region-aware Machine Translation, a type of style-targeted translation. The dataset consists of professional translations from English into two regional variants each of Portuguese and Mandarin Chinese. Source documents are selected to enable detailed analysis of phenomena of interest, including lexically distinct terms and distractor terms. We explore automatic evaluation metrics for FRMT and validate their correlation with expert human evaluation across both region-matched and mismatched rating scenarios. Finally, we present a number of baseline models for this task, and offer guidelines for how researchers can train, evaluate, and compare their own models. Our dataset and evaluation code are publicly available: \url{\frmturl}.
\end{abstract}

\input{1-introduction}

\input{2-rel_work}

\input{3-dataset}

\input{4-metrics}

\input{5-models}

\input{6-results}

\input{7-conclusion}

\section*{Acknowledgments}\label{sec:ack}
For helpful discussion and comments, we thank Jacob Eisenstein, Noah Fiedel, Macduff Hughes and Mingfei Lau. For feedback around regional differences, we thank Andre Araujo, Chung-Ching Chang, Andreia Cunha, Filipe Gonçalves, Nuno Guerreiro, Mandy Guo, Luis Miranda, Vitor Rodrigues, and Linting Xue.

\bibliography{paper.bib, anthology.bib}
\bibliographystyle{acl_natbib}
\end{document}

%% file: 0-preamble.tex
\usepackage{times,latexsym}
\usepackage{url}
\usepackage[T1]{fontenc}

\usepackage[acceptedWithA]{tacl2021v1}

\AddToShipoutPicture{%
\AtPageLowishCenter{\thepage}
}

\usepackage{xspace}

\newif\iftaclinstructions
\taclinstructionsfalse 
\iftaclinstructions
\renewcommand{\confidential}{}
\renewcommand{\anonsubtext}{(No author info supplied here, for consistency with
TACL-submission anonymization requirements)}
\newcommand{\instr}
\fi

\usepackage{multirow}

\usepackage{CJKutf8} 
\newcommand{\simp}[1]{\begin{CJK*}{UTF8}{gbsn}#1\end{CJK*}}
\newcommand{\trad}[1]{\begin{CJK*}{UTF8}{bsmi}#1\end{CJK*}}

\newcommand{\error}[1]{\textcolor{red!60!black}{\underline{#1}}}

\newcommand{\minus}{\scalebox{0.7}[1.0]{$-$}} 

\newcommand{\frmturl}[0]{https://bit.ly/frmt-task}

\usepackage{graphicx}

\usepackage{booktabs}

\title{FRMT: A Benchmark for Few-Shot Region-Aware Machine Translation}

\author{
 Parker Riley\Thanks{Equal contribution.} , Timothy Dozat\footnotemark[1] , Jan A. Botha\footnotemark[1] , Xavier Garcia\footnotemark[1] , \\ {\bf Dan Garrette, Jason Riesa, Orhan Firat, Noah Constant} \\
 Google Research \\
 {\texttt \{prkriley, tdozat, jabot, xgarcia, dhgarrette, riesa, orhanf, nconstant\}@google.com}
}

\date{}

%% file: 1-introduction.tex
\section{Introduction}

Machine translation (MT) has made rapid advances in recent years, achieving impressive performance for many language pairs,
especially those with high amounts of parallel data available.
Although the MT task is typically specified at the coarse level of a language (e.g.~Spanish or Hindi),
some prior work has explored finer-grained distinctions, such as between regional varieties of Arabic \cite{zbib-etal-2012-machine}, or specific levels of politeness in German \cite{sennrich-etal-2016-controlling}.
Unfortunately, most approaches to style-targeted translation thus far rely on large, labeled training corpora \cite{zbib-etal-2012-machine, lakew-etal-2018-neural, costa-jussa-etal-2018-neural, honnet-etal-2018-machine, sajjad-etal-2020-arabench, Wan_Yang_Wong_Chao_Du_Ao_2020, kumar-etal-2021-machine}, and in many cases these resources are unavailable or expensive to create.

\begin{figure}
    \centering
    \includegraphics[width=\columnwidth]{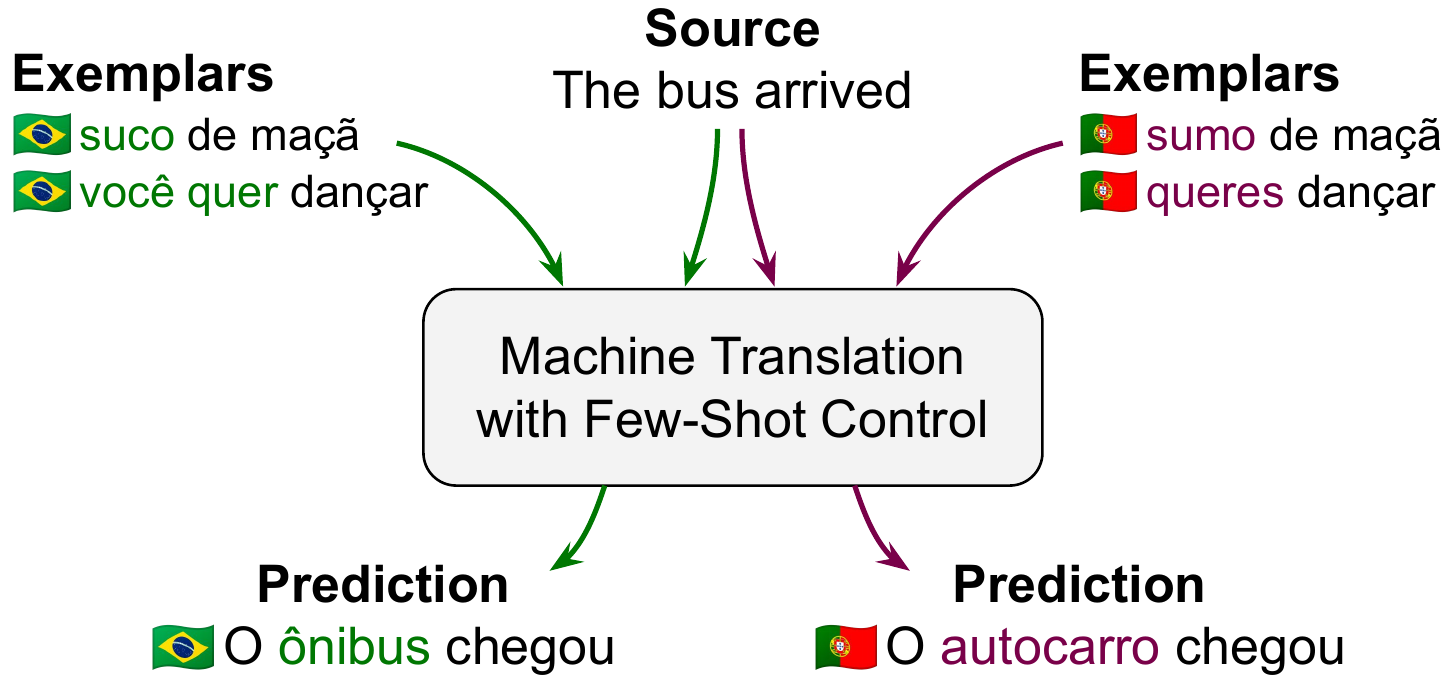} 
    \caption{FRMT requires a machine translation model to adapt its output to be appropriate for a specific region, such as Brazil (left) or Portugal (right). Because only a few exemplars are provided to convey the target region, methods that perform well on FRMT can likely extend to other regions and styles.}
    \label{fig:frmt_diagram}
\end{figure}

We explore a setting for MT where unlabeled training data is plentiful for the desired language pair, but only a few parallel examples (0--100, called ``exemplars'') are annotated for the target varieties. As a specific use-case, we examine translation into regional varieties: Brazilian vs.~European Portuguese and Mainland vs.~Taiwan Mandarin. While these varieties are mutually intelligible, they often exhibit lexical, syntactic, or orthographic differences that can negatively impact an MT user's experience. Figure~\ref{fig:frmt_diagram} illustrates the use of exemplars to control the regional variety at inference time.

MT systems that do not support region or style distinctions may be biased toward varieties with more available data (the ``web-majority'' varieties). 
We observe this bias in a widely used proprietary MT system, with measurable negative effects for speakers of web-minority varieties (\S\ref{sec:results_auto}). One barrier to further research on this issue is the lack of a high-quality evaluation benchmark. Thus, to encourage more access to language technologies for speakers of web-minority varieties and more equitable NLP research, we make the following contributions:
(1)~We construct and release FRMT, a new dataset for evaluating few-shot region-aware translation from English to Brazilian/European Portuguese and Mainland/Taiwan Mandarin.
(2)~We evaluate predictions from a number of existing and custom-trained baseline systems on the FRMT task using automatic metrics.
(3)~We conduct detailed human evaluations of gold and model-based translations on FRMT, under all combinations of rater region and target region.
(4)~We analyze the correlation of automatic metrics and human evaluations on FRMT, and propose a new targeted metric for lexical accuracy.

%% file: 2-rel_work.tex
\section{Related Work}

Textual style transfer aims to control fine-grained stylistic features of generated text. Earlier work leverages supervised parallel data \cite{jhamtani-etal-2017-shakespearizing}; later work assumes labeled but non-parallel training data \cite{shen17style, li-etal-2018-delete, niu18multitask}, or foregoes training-time labels entirely, as in our setting, relying only on few-shot exemplars provided at inference time \cite{xu2020variational, riley-etal-2021-textsettr, garcia2021towards}. However, style transfer evaluation protocols are known to be lacking \cite{pang-gimpel-2019-unsupervised, briakou-etal-2021-review, hu2022tst}, due to the underspecification of stylistic attributes (e.g.~formality, sentiment) and the absence of standardization across studies. Region-aware translation addresses these issues, providing a test-bed for exploring few-shot attribute control---MT evaluation methods are relatively mature, and many regional language varieties can be sufficiently delineated for the task.

Previous work has explored many sub-types of variety-targeted MT\@. Region-aware MT targets specific regions or dialects \cite{zbib-etal-2012-machine, costa-jussa-etal-2018-neural, honnet-etal-2018-machine, lakew-etal-2018-neural, sajjad-etal-2020-arabench, Wan_Yang_Wong_Chao_Du_Ao_2020, kumar-etal-2021-machine}; formality-aware MT targets different formality levels \cite{niu-etal-2017-study, niu-etal-2018-multi, wang-etal-2019-harnessing}; and personalized MT aims to match an individual's specific style \cite{michel-neubig-2018-extreme, vincent-2021-towards}. However, with few exceptions (e.g.~\citealt{garcia2021towards}), these works assume the availability of large-scale datasets containing examples with the target varieties explicitly labeled. In the present work, we design a benchmark that emphasizes few-shot adaptability. Although our dataset is limited to four regions and two languages, the few-shot setup and high degree of linguistic dissimilarity between the selected languages means that approaches performing well on the entire FRMT benchmark can be expected to generalize reasonably well to other languages, other regions, and other stylistic attributes.

Several existing parallel corpora cover regional language varieties, but have limitations that motivate us to construct a new high-quality, targeted dataset. e-PACT \cite{barreiro2017pact} comprises translations from English books into Portuguese variants, but is small and not easily accessible. OpenSubTitles \cite{lison-etal-2018-opensubtitles2018} skews toward shorter utterances and is noisy due to automatic alignment. WIT3 \cite{cettolo-etal-2012-wit3} provides translations of TED-talk transcripts into many languages, but relies on volunteer translators which may limit quality.

Popular shared tasks have not included region-targeted translation either:
The Conference on Machine Translation (WMT) has included translation between similar languages \cite[e.g.][]{akhbardeh-etal-2021-findings}, while the Workshop on NLP for Similar Languages, Varieties and Dialects (VarDial) focuses mainly on classification and not translation \cite[e.g.][]{vardial-2021-nlp}.

Furthermore, we are not aware of previous work that (1) measures deltas in human evaluation metrics between the region-matched and region-mismatched settings, (2) correlates these with automated metrics, (3) offers tailored sub-tasks targeting region-differentiated lexical items and region-biased distractors, or (4) defines targeted metrics testing region-appropriateness.

%% file: 3-dataset.tex
\section{FRMT Dataset}\label{sec:new-dataset}

We introduce the FRMT dataset for evaluating the quality of few-shot region-aware machine translation. The dataset covers two regions each for Portuguese (Brazil and Portugal) and Mandarin (Mainland and Taiwan). These languages and varieties were selected for multiple reasons: (1) They have many speakers who can benefit from increased regional support in NLP. (2) Portuguese and Mandarin are linguistically very distinct, coming from different families; we therefore hypothesize that methods that perform well on both are more likely to generalize well to other languages.
The dataset was created by sampling English sentences from Wikipedia and acquiring professional human translations in the target regional varieties.
Final quality verification is done through manual evaluation by an independent set of translators, using the MQM protocol \cite{Freitag2021ExpertsEA} that we also employ to evaluate system translation quality.

\subsection{Data sampling method}\label{sec:data_sampling}
FRMT seeks to capture region-specific linguistic differences, as well as potential distractors. To this end, we divide the dataset into three buckets (\texttt{lexical}, \texttt{entity}, \texttt{random}), each containing human translations of sentences extracted from different sets of English Wikipedia articles.\footnote{%
As Wikipedia data source we use the training split of {\tt wiki40b} (v1.3.0) by \citet{wiki40b}, available at \scriptsize{\url{https://www.tensorflow.org/datasets/catalog/wiki40b}}.}

{\bf Lexical}: We collect English lexical items for which the best translation into the target language differs depending on the target region. To source these, we rely on blogs and educational websites that list terms differing by region. We further validate each pair of translations by asking a native speaker of each region whether each translation is appropriate for the intended meaning in their region. We filter to only use pairs where exactly one translation is appropriate per region. This is done independently for Portuguese and Mandarin as target languages, yielding lists of 23 and 15 terms, respectively. For each term $t$, we extract up to 100 sentences from the beginning of the English Wikipedia article with title $t$.

{\bf Entity}:  We select entities that are strongly associated with specific regions under consideration (e.g.~Lisbon and S\~{a}o Paulo), which may have adversarial effects for models that rely heavily on correlations learned from pretraining.
Our selection comprises 38 Mandarin-focused and 34 Portuguese-focused entities.
We extract up to 100 source sentences from the beginning of the English Wikipedia article about each selected entity.

{\bf Random}: 
For a more naturally-distributed subset, we randomly sample 100 articles from Wikipedia's collections of ``featured'' or ``good'' articles.\footnote{%
\scriptsize{\url{https://en.wikipedia.org/wiki/Wikipedia:Good_articles/all}} and
\scriptsize{\url{https://en.wikipedia.org/wiki/Wikipedia:Featured_articles}} (as of 2021-12-15).
}
Here, we take up to 20 sentences from the start of a randomly chosen section within each article.
Unlike the other two buckets, this one features one common set of sentences to be translated into all four target variants.

\subsection{Human translation}
14 paid professionals translated the selected English texts into the four target language variants: 3 translators per Portuguese region and 4 per Mandarin region. For each region, each sentence was translated by one translator, resulting in one reference per source. Each translator translated non-overlapping chunks of the source data one sentence at a time in the order of the original text.
Sentences that were rejected by at least one translator (e.g.~for having too much non-English text) are not included in our dataset.

\subsection{Corpus statistics}
For each bucket, we split our data into exemplar, development (dev), and test data. The exemplars are intended to be the only pairs where the region label is shown to the model, such as via few-shot or in-context learning \cite{brown-etal-22-gpt3}. Providing these ensures increased comparability across methods on the FRMT benchmark, in addition to sidestepping potential domain mismatch issues by providing exemplars from the same domain (Wikipedia text) as the evaluation data.

Table~\ref{tab:corpus_stats} reports the number of released sentence pairs for each split of the dataset. Sentences from a given Wikipedia page appear only in a single split, ensuring a system cannot ``cheat'' by memorizing word--region associations from the exemplars, or by overfitting to words and entities while hill-climbing on the dev set.

Table~\ref{tab:dataset-examples} shows example items from each bucket.

\begin{table}
\centering
\footnotesize
\begin{tabular}{ccrr}
\toprule
\textbf{Bucket} & \textbf{Split} & \textbf{Portuguese} & \textbf{Mandarin} \\
\midrule
\multirow{3}{3.4em}{Lexical} & Exemplar & 118 & 173\\
& Dev & 848 & 524\\ 
& Test & 874 & 538\\ 
\midrule
\multirow{3}{3.4em}{Entity} & Exemplar & 112 & 104\\
& Dev & 935 & 883\\ 
& Test & 985 & 932\\ 
\midrule
\multirow{3}{3.4em}{Random} & Exemplar & 111 & 111\\
& Dev & 744 & 744\\ 
& Test & 757 & 757\\ 
\midrule
\multirow{3}{3.4em}{Total} & Exemplar & 341 & 388\\
& Dev & 2527 & 2151\\ 
& Test & 2616 & 2227\\ 
\bottomrule
\end{tabular}
\caption{Number of sentence pairs by bucket, split, and language, as well as cross-bucket totals. Note, the \texttt{random} bucket contains the same English source sentences across the Portuguese and Mandarin sets.}\label{tab:corpus_stats}
\end{table}

\begin{table*}[t]
\centering
\footnotesize
\begin{tabular}{lp{0.4\textwidth}p{0.4\textwidth}}
\toprule
\textbf{Bucket} &
  \textbf{pt-BR} &
  \textbf{pt-PT} \\ \midrule
\multirow{2}{*}{\textbf{lexical}}
 & 
  Em 2019, a Virgin Atlantic começou a permitir que suas comissárias de bordo femininas usassem calças e não usassem maquiagem. &
  Em 2019, a Virgin Atlantic começou a autorizar as assistentes de bordo a usar calças e a dispensar maquilhagem. \\[1ex]
  & \multicolumn{2}{p{0.8\textwidth}}{\textit{In 2019, Virgin Atlantic began to allow its female flight attendants to wear pants and not wear makeup.}} \\
  \midrule
\multirow{2}{*}{\textbf{entity}} 
 &
  Os ônibus são o meio mais barato de se movimentar por Natal. &
  Os autocarros são a maneira mais barata de viajar pelas localidades próximas de Natal. \\ [1ex]
 &
  \multicolumn{2}{p{0.8\textwidth}}{\textit{Buses are the cheapest way to move around Natal.}} \\
  \midrule
\multirow{2}{*}{\textbf{random}} 
 &
  O suco causa alucinações psicodélicas intensas em quem o bebe, e a polícia logo o rastreou até a fazenda e partiu para prender Homer, Seth e Munchie. &
  O sumo provoca fortes alucinações psicadélicas a quem bebe do mesmo e a polícia rapidamente segue o rasto até à quinta, deslocando-se até lá para prender Homer, Seth e Munchie. \\[1ex]
  &
  \multicolumn{2}{p{0.8\textwidth}}{\textit{The juice causes intense psychedelic hallucinations in those who drink it, and the police quickly trace it to the farm and move in to arrest Homer, Seth, and Munchie.}} \\
  \bottomrule
\end{tabular}
\caption{Examples from the dataset, limited to the Portuguese dev-set for brevity. The last two columns show the reference human translations obtained for each region given the English source text (in italics).
For the lexical and entity buckets, we show examples for which the Levenshtein edit-distance between the two translations is near the median observed for the whole dev-set.
}
\label{tab:dataset-examples}
\end{table*}

\subsection{Limitations}\label{sec:limitations}

Our dataset is designed to capture differences in regional varieties, but capturing all such differences in a finite dataset is impossible. While we specifically target lexical differences, the terms were selected via a manual process based on online resources that discuss lexical differences in these languages, and these resources can sometimes be incorrect, outdated, or inattentive to rare words or words with more subtle differences. Other regional distinctions, such as grammatical differences, were not specifically targeted by our data bucketing process, and thus the degree to which they are captured by the dataset is determined by their likelihood to occur in translations of English Wikipedia text. This also means that differences that only surface in informal settings are unlikely to be included, as Wikipedia text has a generally formal style.

While we believe that methods that perform well on all four varieties included in FRMT should be applicable to other languages and varieties, measuring this would require a similar dataset with wider coverage. Constructing such a dataset requires only knowledge of regional differences to inform selection of source texts as in our \texttt{lexical} and \texttt{entity} buckets, and translators who are native speakers of the target varieties. An additional pool of MQM-trained translators would be needed to validate the collected translations for regional distinctiveness.

In spite of validation through MQM, it should be noted that the region-targeted translations we collected are not necessarily minimal contrastive pairs, but may include differences arising from factors other than regional variation, such as individual style preferences of the human translators.

%% file: 4-metrics.tex
\section{Evaluation Metrics}\label{sec:metrics}

While human judgments are the gold standard for evaluating machine-generated texts, collecting them can be time-consuming and expensive. For faster iteration, it can be helpful to measure progress against automatic metrics that are known to correlate well with human judgments. We hypothesize that common reference-based MT evaluation metrics might have differing sensitivities to regional differences, and so we conduct a human evaluation of several baseline models (see \S\ref{sec:results_human}) and compute correlation of several automatic metrics with the human judgments. We also propose a new automated lexical accuracy metric that more directly targets region-awareness.

\subsection{Human evaluation}\label{sec:human_eval_mqm}
To obtain the highest fidelity human ratings, we use the expert-based 
Multidimensional Quality Metrics (MQM) evaluation framework proposed by \citet{Freitag2021ExpertsEA} and recommended by the WMT'21 Evaluation Campaign \citep{freitag-etal-2021-results}. We show expert raters chunks of 10 contiguous English sentences from our test set with one corresponding set of translations. Raters then identify errors in the translations, assigning a category and severity to each. Due to cost constraints, we evaluate 25\%{} of the test set, evenly distributed across our three evaluation buckets. Within each region, each chunk is rated by three raters, who achieve interannotator consistency of $70.4\pm2.2$ (as 100-scaled intraclass correlation\footnote{Using the \texttt{icc} function of R's \texttt{irr} library \citep{gamer2019irr}.}).

Each output is shown to raters of \emph{both} regions of the corresponding language. 
All Mandarin outputs are automatically transliterated into the rater's region's corresponding Han script (Mainland: simplified; Taiwan: traditional), using Google Translate ``Chinese (Simplified)'' $\leftrightarrow$ ``Chinese (Traditional)''.

\subsection{Automatic translation quality metrics}\label{sec:translate_quality_metrics}
We evaluate the following automatic, reference-based metrics:

\textbf{BLEU} \citep{papineni-etal-2002-bleu}: Based on token $n$-grams, using {\small \texttt{corpus\_bleu}} from \citet{post-2018-call}.\footnote{SacreBLEU version strings for \{Portuguese,Mandarin\}: {\scriptsize BLEU|nrefs:1|case:mixed|eff:no|tok:\{13a,zh\}|smooth:exp|version:2.3.1 }}

\textbf{\textsc{chrF}} \citep{popovic-2015-chrf}: Based on character $n$-gram F1, using {\small \texttt{corpus\_chrf}} from \citet{post-2018-call}.\footnote{SacreBLEU version string:\\ {\scriptsize chrF2|nrefs:1|case:mixed|eff:yes|nc:6|nw:0|space:no|version:2.3.1}}

\textbf{BLEURT} \citep{sellam-etal-2020-bleurt}: A learned, model-based metric, that has good correlation with human judgments of translation quality. To the best of our knowledge, BLEURT has not been evaluated with respect to human judgments of \textit{region-specific} translation quality.

\textbf{BLEURT-D\{3,6,12\}} \citep{sellam-etal-2020-bleurt}: These distilled versions of BLEURT are less resource-intensive to run, and have 3, 6, and 12 layers, respectively.
For all BLEURT variants, we use checkpoints released by its authors.

As in the human evaluation, all Mandarin outputs are transliterated into the target regional Han script before evaluation.

\subsection{Correlation}
For computing correlation, each data point is a score on a 10-sentence chunk of model output, covering the three models discussed in section \S\ref{sec:results_human}, using both matched and mismatched ratings.
For MQM, this is the average of 30 weighted ratings: one per sentence per rater. The category/severity weights are described in \citet{Freitag2021ExpertsEA}.
For BLEU and \textsc{chrF}, which are corpus-level metrics, we take the 10 input/output sentence pairs as the ``corpus''. For BLEURT, we use the average sentence-level score. 
Table \ref{tab:correlation} presents the correlation results, scaled by $\minus{}$100.\footnote{We negate the correlations with MQM because higher quality corresponds to lower MQM scores.}

\begin{table}
    \centering
    \footnotesize
    \begin{tabular}{lcc}
         \toprule
         \bf Metric & \bf Kendall's $\tau$ & \bf Pearson's $\rho$\\\midrule
         \textsc{chrF} & 43.6 & 48.4\\
         \textsc{BLEU} & 44.9 & 57.5\\
         \textsc{BLEURT-D3} & 50.6 & 63.1\\
         \textsc{BLEURT-D6} & 50.7 & 63.3\\
         \textsc{BLEURT-D12} & 51.2 & 64.0\\
         \textsc{BLEURT} & 52.4 & 65.4 \\
         \bottomrule
    \end{tabular}
    \caption{Coefficients of correlation between human MQM ratings and several automated metrics. \textsc{chrF} has the lowest correlation, with \textsc{BLEU} performing slightly better. All \textsc{BLEURT} models outperform the non-learned metrics, with the full-size model achieving higher correlation than the smaller distillations thereof.}
    \label{tab:correlation}
\end{table}

We observe that the learned BLEURT metrics outperform the non-learned metrics by wide margins, in line with findings from \citet{sellam-etal-2020-bleurt} that neural methods outperform $n$-gram based methods. 
Additionally, the teacher model (BLEURT) outperforms the distilled student models, with larger students consistently outperforming smaller ones. 

\subsection{Lexical accuracy}\label{sec:lexical_accuracy}

To assess a model's ability to select lexical forms appropriate to the target region, we define a \emph{lexical accuracy} metric.  As discussed in section \S\ref{sec:data_sampling}, sentences in the \texttt{lexical} bucket are from Wikipedia articles containing specific words that we expect to have distinct regional translations.
For instance, we include source sentences from the English Wikipedia article ``Bus'' in the Portuguese \texttt{lexical} bucket, as the word for bus is distinct in Brazil and Portugal (\textit{ônibus} vs.~\textit{autocarro}).
As the expected output words are known ahead of time, we can directly measure the rate at which a model selects region-appropriate variants.

Starting from the list of terms used to select articles for the \texttt{lexical} bucket, we remove the terms selected for the exemplars split in order to test generalization to unseen terms. This results in 18 term-pairs in Portuguese and 13 in Mandarin.

We calculate the metric over all model outputs for the \texttt{lexical} bucket, covering both regions. For each term-pair, we calculate the number of sentences containing the matched variant and the number of sentences containing the mismatched variant.
The model's lexical accuracy (LA) for the given language is then the total number of matches divided by the sum of matches and mismatches:

\begin{equation}
\textrm{LA} = \frac{N_{match}}{N_{match} + N_{mismatch}}
\end{equation}

To account for Portuguese inflection, we considered matching lemmatized forms rather than surface forms, but found little difference in the resulting scores. We thus report results using naive surface matching, which avoids a dependency on a specific lemmatizer and improves reproducibility.

To disentangle lexical choice from script choice, we define lexical accuracy to be script-agnostic---e.g., for the word \emph{pineapple}, if the target is zh-TW, we count both script forms of the Taiwan variant \textit{fènglí} (\trad{鳳梨} and \simp{凤梨}) as correct, and both script forms of the Mainland variant \textit{bōluó} (\simp{菠萝} and \trad{菠蘿}) as incorrect. This ensures that models are judged solely on their lexical choices, and prevents ``gaming'' the metric by only using the lexical forms and script of a single region.

We emphasize that lexical choice is just one important facet of region-aware translation, aside from morphology, syntax, and beyond.
Even so, we believe that this easy-to-calculate metric is worth iterating on, since one may safely say that a model that scores poorly on lexical accuracy has not solved region-aware translation.

\subsection{Reporting FRMT results}\label{sec:recommendations}

For the FRMT \emph{task} (as opposed to the \emph{dataset}), we stipulate a key ``few-shot'' restriction: candidate models \textbf{may not be intentionally exposed to any region-labeled data} at any point during training, except for data from the FRMT exemplars split.  This restriction covers both region-labeled monolingual data as well as region-labeled parallel translation data.\footnote{Models may train on multilingual web crawl data, as is common practice, as long as supervised region labels are not provided.  We allow that some implicit or explicit region labels may appear by chance within the unsupervised data.} While it may not be difficult to obtain region labels for Brazil/Portugal or Mainland/Taiwan (e.g.~by filtering web pages on top-level web domain), we intend for FRMT to serve as a measure of few-shot generalization to \emph{arbitrary} regions and language varieties, for which obtaining labels may be much harder.

Researchers sharing FRMT results should \textbf{report lexical accuracy, per-bucket BLEU, and the ``FRMT'' score} (described in \S\ref{sec:results_auto}) on test, as shown in Tables \ref{tab:results_auto} and \ref{tab:lexical_accuracy}. These metrics can be calculated with our provided evaluation scripts.\footnote{Scripts available at \url{\frmturl}.}

We also recommend reporting BLEURT scores, but recognize that this may not always be possible, as it requires significantly more computational resources.  Similarly, we encourage human evaluation using MQM as a gold standard, but do not wish to promote this as a community metric, due to its impracticality for many researchers and the potential confound of having different rater pools.

Finally, for any model candidate, it is important to \textbf{report how many exemplars were supplied} for each variety. To improve comparability, we recommend 0, 10, or 100 exemplars per region.

%% file: 5-models.tex
\section{Baseline Models}\label{sec:models}

\begin{figure*}
    \centering
    \includegraphics[width=0.45\textwidth]{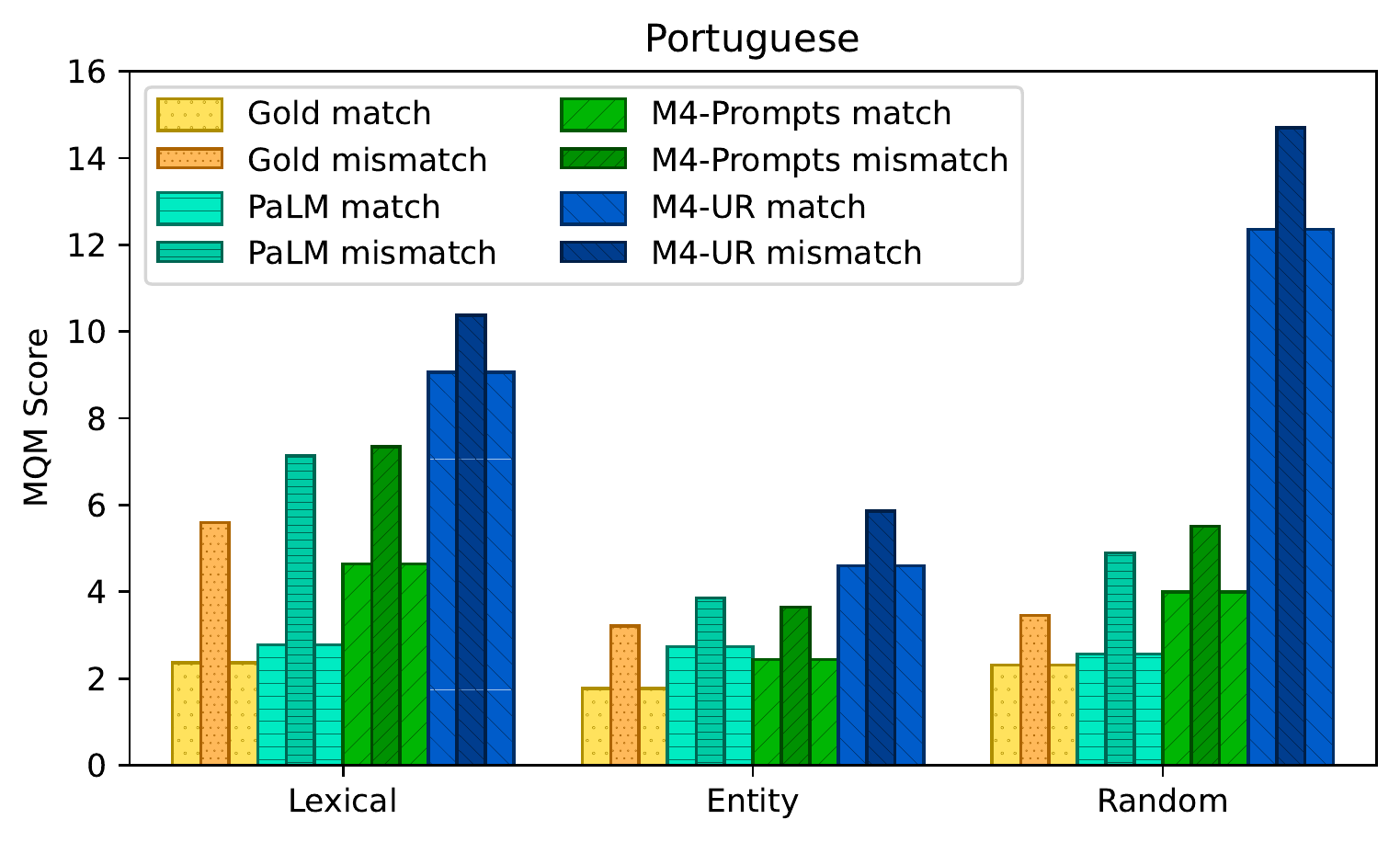} 
    \includegraphics[width=0.45\textwidth]{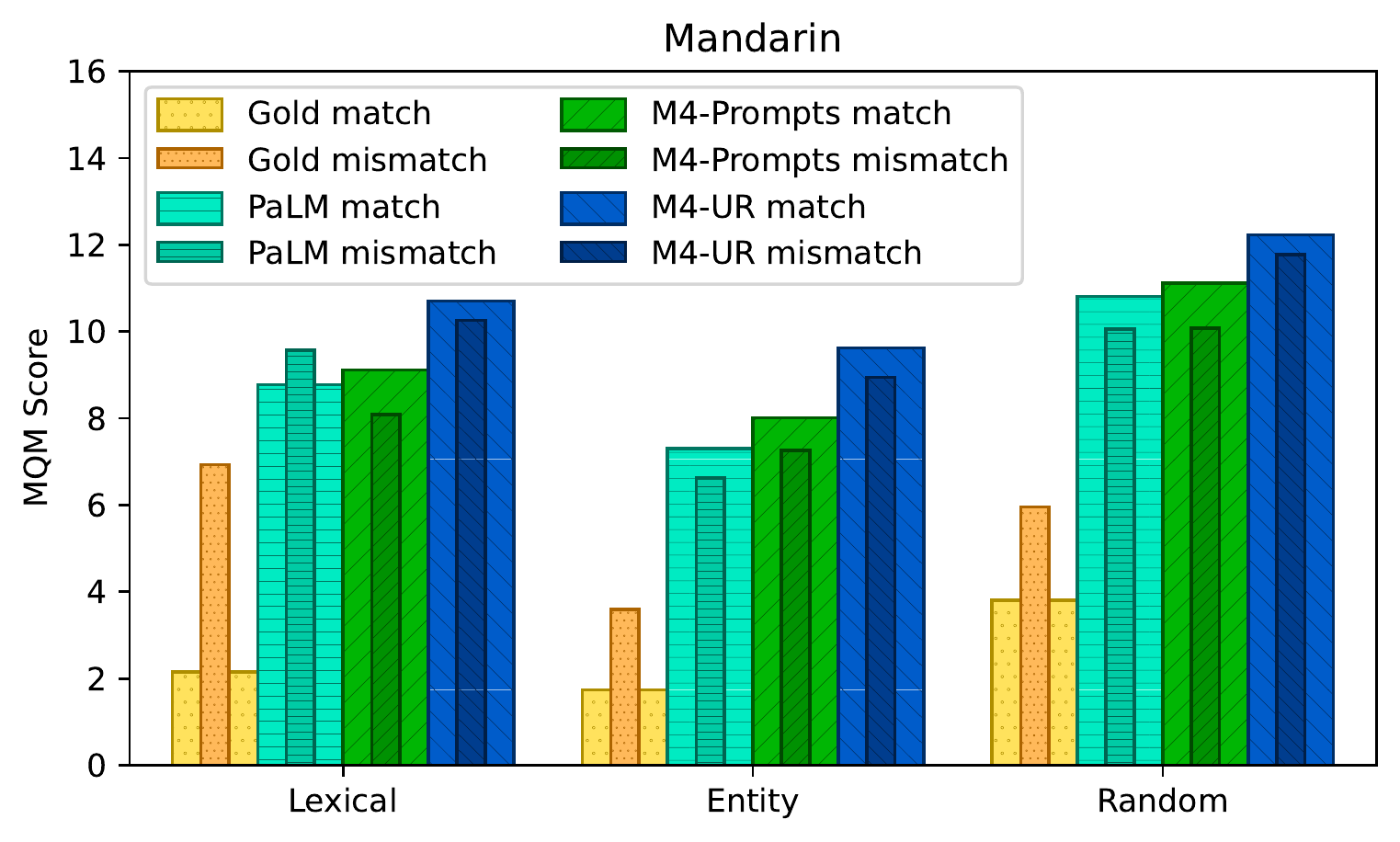}
    \caption{MQM ($\downarrow$) scores for gold translations and model predictions in Portuguese (left) and Mandarin (right). Thick ``match'' bars show scores from raters in the target region. Thin ``mismatch'' bars show scores from raters in the opposite region. In all conditions, raters prefer region-matched gold translations, confirming the presence of region-specific phenomena in the collected data. PaLM is the highest-rated baseline, but still has room for improvement, particularly in Mandarin.}
    \label{fig:mqm_scores}
\end{figure*}

We evaluate a handful of academic MT models that claim some ability to provide few-shot controllable translations. We also evaluate a commercial MT system that does not distinguish between these regional varieties.

Our first baseline is the Universal Rewriter (\textbf{UR}) of \citet{garcia2021towards}, which supports multilingual style transfer and translation. It is initialized from an mT5-XL checkpoint \citep{xue-etal-2021-mt5} and finetuned on a combination of monolingual and parallel data from mC4 and OPUS, respectively. We train it with sequence length of $128$ instead of $64$, to be directly comparable to our other models.

Our second baseline is UR finetuned from the Massively Multilingual Massive Machine translation (M4) model of \citet{siddhant-etal-22-towards} instead of mT5 (\textbf{M4-UR}). We hypothesize that initializing from a model explicitly designed for translation will outperform one trained as a general language model. For both UR and M4-UR, we use the first 100 exemplars from the \texttt{lexical} buckets.

Our third baseline uses natural language prompting to control the regional variety of M4's output (\textbf{M4-Prompts}), such as prefixing the input with ``A Brazilian would write it like this:''. This is motivated by earlier work using this technique effectively for large language models \citep{wei2022finetuned, sanh2022multitask, brown-etal-22-gpt3}, and more recent work applying it to region-aware MT \citep{garcia2022using}.

Our fourth baseline fine-tunes the M4-Prompts model, where the source-side language tags used to induce the target language are replaced with prompts of the form ``Translate to [language]:''. This model (\textbf{M4-Prompts FT}) is designed to explicitly introduce prompting behavior. At inference time, we replace ``[language]'' with the variety name (e.g.~``Brazilian Portuguese''). Neither M4-Prompts nor M4-Prompts FT use exemplars.

Our next three baselines are different-sized versions of PaLM \citep{chowdhery2022palm}, a large language model that has demonstrated remarkable zero-shot and few-shot performance on a variety of tasks (\textbf{PaLM 540B}, \textbf{PaLM 62B}, and \textbf{PaLM 8B}, referring to their approximate parameter counts).
The prompt for these models begins with ``Translate the following texts from English to [language variety]'' and is followed by ten exemplars selected randomly from the \texttt{lexical} bucket.\footnote{The model has a fixed input sequence length, including the prompt, and a fixed output sequence length. We ensure that the ten exemplars are short enough to leave at least 128 tokens for the input text, to match the 128 tokens allotted to the output.} Each exemplar is put on two lines: first the English text, prefixed by ``English:'', and then the translation in the target variety, prefixed by the variety's name. At the end of the prompt, we show the model the input text and the language variety prefix, and take the first decoded line of text.

Finally, we examine \textbf{Google Translate}\footnote{\url{translate.google.com}, accessed April 4th, 2022}, a publicly-available commercial MT model that does not support regional varieties for Portuguese or Mandarin (though it does support transliteration of traditional and simplified scripts). We evaluate this system mainly to test the hypothesis that variety-agnostic systems will be biased toward the web-majority variety.

%% file: 6-results.tex
\section{Baseline Model Performance}\label{sec:results}

\subsection{Human evaluation results}\label{sec:results_human}

We select three baseline models for human evaluation: M4-UR, M4-Prompts, and \mbox{PaLM 540B}\@, covering a variety of modeling techniques.

Figure \ref{fig:mqm_scores} presents human evaluation of our baselines on the 25\% sample of our test set described in \S\ref{sec:translate_quality_metrics}. 
For the gold data, we observe that raters of all regions prefer translations from their own region (the ``matched'' case) over translations from the other region (the ``mismatched'' case) in all three buckets; when averaged over buckets, the MQM penalties for the matched and mismatched cases are significantly different ($1.73$ matched and $3.55$ mismatched; $t=-3.34; p<0.001$). This indicates that, despite the limitations discussed in \S\ref{sec:limitations}, our data collection process succeeded in producing regionally-distinct translations.
This effect is strongest in the \texttt{lexical} bucket, presumably due to the high rate of region-distinct terms in these sentences.

In Portuguese, we find that all models perform better in the region-matched setting, indicating that each model has some ability to localize to Brazil and Portugal.  However, in Mandarin, apart from PaLM's \texttt{lexical} bucket, region match does not lead to MQM gains, indicating that these models are not able to produce better, more region-specific translations in this case.

Comparing across models, we find that PaLM performs the best, followed by M4-Prompts and then M4-UR, consistently across both Portuguese and Mandarin. PaLM performs particularly well in the \texttt{lexical} bucket, suggesting that larger models may be better suited to the task of memorizing region-specific lexical variants.

For Mandarin, a large gap remains between expert translations and our baselines: averaged over buckets, the gold matched MQM penalty is $2.5$ vs. PaLM's $8.8$. It's apparent that better region handling will be needed to close this gap, since our baselines have much worse match/mismatch deltas than gold translations: the average gold mismatched penalty minus matched penalty was $2.7$, while PaLM's was $-0.3$.

For Portuguese, while PaLM gives impressive results, there is still a meaningful gap with expert translation: averaged over buckets, the gold MQM penalty was $2.1$ vs. PaLM's $2.7$, indicating headroom for our task. There is also the important question of whether competitive performance can be achieved with smaller models, which are better suited for production use-cases.

\begin{figure}
    \centering
    \includegraphics[width=\columnwidth]{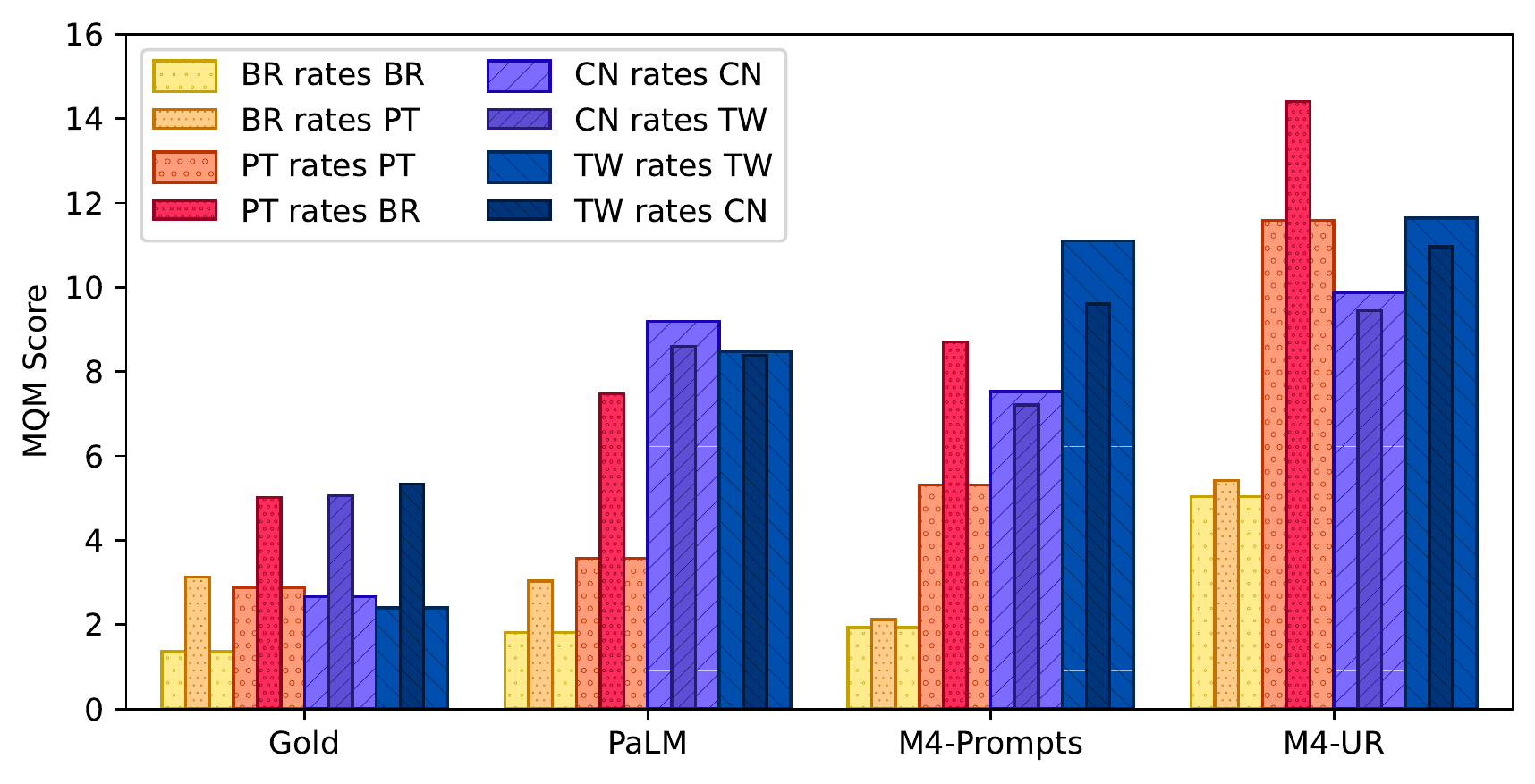} 
    \caption{MQM ($\downarrow$) scores for gold translations and model predictions, broken down by rater region and target region. For example ``BR rates PT'' indicates Brazilian raters scoring sentences targeted to Portugal.}
    \label{fig:mqm_scores_by_rater}
\end{figure}

Figure \ref{fig:mqm_scores_by_rater} breaks down scores by rater and target region, over the full 25\% sample. As before, in each setting, raters prefer region-matched over mismatched gold translations. For Portuguese, we find that our pt-PT raters were ``harder graders'' than our pt-BR raters, with a delta of +$2$ MQM between the regions in both matched and mismatched settings; by contrast, our Mandarin raters were well calibrated across regions.

We further examined model performance on the \texttt{entity} bucket, to test whether the presence of ``distractor'' entities (associated with the non-target region) would hurt translation quality, but we did not find significant differences in MQM scores. Still, we note isolated examples of this effect; for instance, when targeting pt-BR, the M4-Prompts model produces the pt-PT spelling \textit{patrim\'onio} (cf.~pt-BR \textit{patrim\^onio}), but only when the English source contains the words \textit{Lisbon} or \textit{Portugal}. We expect the \texttt{entity} bucket will be useful to researchers looking for similar effects.

\subsection{Automated metric results}\label{sec:results_auto}

\begin{table*}
\centering
\footnotesize
\begin{tabular}{lcccccc|c}
\toprule
& \multicolumn{2}{c}{\bf Lexical} & \multicolumn{2}{c}{\bf Entity} & \multicolumn{2}{c}{\bf Random} & \bf FRMT \\
\bf Model & pt-BR & pt-PT & pt-BR & pt-PT & pt-BR & pt-PT & pt\\\midrule
UR & 37.4 (69.9) & 32.7 (68.0) & 46.7 (76.3) & 40.8 (73.6) & 39.8 (70.7) & 35.3 (69.2) & 38.7 (71.3)\\
M4-UR & 46.7 (74.5) & 32.7 (69.7) & 53.5 (79.9) & 45.4 (77.5) & 43.1 (70.9) & 32.9 (68.4) & 42.0 (73.5)\\
M4-Prompts & 54.1 (77.1) & 36.9 (72.1) & 56.9 (81.1) & 47.3 (78.4) & 56.1 (77.5) & 41.0 (73.7) & 48.2 (76.6)\\
M4-Prompts FT & 45.5 (70.1) & 32.5 (67.4) & 48.6 (73.8) & 40.7 (72.8) & 48.1 (70.5) & 36.9 (69.0) & 41.7 (70.6)\\
PaLM 8B & 38.6 (69.8) & 26.7 (65.8) & 45.9 (75.9) & 38.0 (73.6) & 39.3 (69.4) & 32.1 (67.8) & 36.5 (70.4)\\
PaLM 62B & 49.5 (75.9) & 36.7 (72.4) & 55.4 (80.1) & 46.1 (77.8) & 50.3 (75.2) & 41.5 (73.5) & 46.3 (75.8)\\
PaLM 540B & 53.7 (77.1) & \textbf{40.1} (\textbf{73.9}) & \textbf{59.0} (\textbf{81.2}) & \textbf{49.5} (\textbf{79.0}) & 54.8 (76.9) & \textbf{45.6} (\textbf{75.5}) & \textbf{50.2} (77.3)\\
Google Translate & \textbf{56.2} (\textbf{78.7}) & 35.6 (72.3) & 56.3 (\textbf{81.2}) & 46.9 (78.3) & \textbf{65.2} (\textbf{80.5}) & 42.9 (75.0) & 49.8 (\textbf{77.6}) \\
\midrule
& zh-CN & zh-TW & zh-CN & zh-TW & zh-CN & zh-TW & zh\\\midrule
UR & 22.6 (58.5) & 13.8 (56.0) & 26.7 (67.1) & 19.5 (65.3) & 26.4 (62.1) & 20.4 (61.0) & 21.3 (61.7)\\
M4-UR & 33.3 (65.0) & 18.9 (58.2) & 43.2 (73.0) & 31.4 (70.4) & 40.8 (65.4) & 30.8 (63.6) & 32.5 (65.9)\\
M4-Prompts & 33.3 (64.9) & 18.3 (57.6) & 44.2 (72.5) & 32.0 (68.7) & 43.7 (67.0) & 32.2 (63.4) & 33.3 (65.6)\\
M4-Prompts FT & 33.8 (65.7) & 18.8 (59.0) & 44.8 (73.2) & 31.6 (69.8) & 42.7 (66.7) & 31.5 (64.0) & 33.2 (66.4)\\
PaLM 8B & 17.6 (55.7) & 13.3 (52.3) & 28.1 (65.7) & 24.4 (63.9) & 21.6 (56.3) & 18.2 (56.1) & 20.4 (58.3)\\
PaLM 62B & 29.2 (62.2) & 20.4 (59.8) & 40.2 (71.8) & 33.0 (69.9) & 34.5 (64.0) & 26.0 (63.1) & 30.3 (65.1)\\
PaLM 540B & 34.8 (66.5) & \textbf{24.6} (\textbf{63.3}) & 44.9 (74.7) & 35.2 (\textbf{72.5}) & 40.0 (67.8) & 29.6 (66.0) & 34.5 (68.4)\\
Google Translate & \textbf{39.7} (\textbf{68.0}) & 21.9 (61.8) & \textbf{50.4} (\textbf{75.0}) & \textbf{37.0} (72.2) & \textbf{56.1} (\textbf{72.0}) & \textbf{39.9} (\textbf{68.7}) & \textbf{40.1} (\textbf{69.6}) \\
 \bottomrule
\end{tabular}
\caption{FRMT per-bucket test set results, in the format: BLEU (BLEURT). The ``FRMT'' score is the geometric mean across regions of the arithmetic mean across buckets.}
\label{tab:results_auto}
\end{table*}

Table \ref{tab:results_auto} shows performance of our baseline models on the automated metrics BLEU and BLEURT\@. \textbf{``FRMT'' score} is a summary of per-language performance, calculated as the geometric mean across regions of the arithmetic mean across buckets.

As mentioned at the outset, we observe that region-agnostic models have a strong bias toward the region with larger presence in web-crawled corpora. This is especially apparent in the \texttt{lexical} bucket, where Google Translate has a +$20.6$ BLEU gap between pt-BR and pt-PT and a +$17.8$ gap between zh-CN and zh-TW\@.

Within the \texttt{lexical} bucket, we note that PaLM outperforms the public Google Translate model in web-minority regions (pt-PT and zh-TW) despite being trained in a fully unsupervised manner. This highlights that even with minimal region-labeled data ($10$ exemplars), it is possible to make meaningful progress over region-agnostic approaches.

\begin{table}
\centering
\footnotesize
\begin{tabular}{lcc}
\toprule
\textbf{Model} & \textbf{pt} & \textbf{zh} \\
\midrule
Gold & 98.6 & 94.4 \\
\midrule
UR & 50.4 & 50.6 \\
M4-UR & 51.2 & 50.9 \\
M4-Prompts & 66.7 & 50.0 \\
M4-Prompts FT & 66.7 & 51.0 \\
PaLM 8B & 85.0 & 69.0 \\
PaLM 62B & 90.4 & 70.8 \\
PaLM 540B & \textbf{93.2} & \textbf{83.6} \\
Google Translate & 50.0 & 50.0 \\
\bottomrule
\end{tabular}
\caption{Lexical accuracy on FRMT test. PaLM outperforms other approaches, while region-agnostic models like Google Translate are guaranteed 50\%.}
\label{tab:lexical_accuracy}
\end{table}

Table \ref{tab:lexical_accuracy} shows \textbf{lexical accuracy} performance, assessing whether specific terms receive region-appropriate translations. Here, the PaLM models outperform alternatives by a wide margin. As even the smallest PaLM model has more than $2\times$ the parameters of our other baselines (3.7B parameters each), this suggests that model capacity is a key ingredient for learning to use region-specific terminology in a few-shot manner. Still, there is a wide gap compared to human performance.

Notably, while the smaller PaLM models outperform our UR and M4 baselines on lexical accuracy, they underperform on BLEU and BLEURT\@. This highlights that using region-appropriate terminology is only a small part of the translation task, and at smaller sizes, models designed specifically for translation have the advantage.

\subsection{Mismatched outputs}\label{mismatched_outputs}

Given a reference in a specified language variety (e.g.~pt-PT), a ``good'' model should achieve a higher score when translating into that variety (the ``matched'' case) than an alternative variety (e.g.~pt-BR; the ``mismatched'' case). To measure the extent to which this holds for our baseline models, we show the delta between matched and mismatched outputs on the test set in Table \ref{tab:mismatched_outputs}.

\begin{table*}
\centering
\footnotesize
\begin{tabular}{lcccccc|c}
\toprule
& \multicolumn{2}{c}{\bf Lexical} & \multicolumn{2}{c}{\bf Entity} & \multicolumn{2}{c}{\bf Random} & \bf $\Delta$FRMT \\
\bf Model & pt-BR & pt-PT & pt-BR & pt-PT & pt-BR & pt-PT & pt\\\midrule
UR & 1.3 (1.0) & \minus{}0.2 (\minus{}0.8) & 1.0 (0.8) & \minus{}1.0 (\minus{}0.8) & 1.5 (0.8) & \minus{}0.7 (\minus{}0.7) & 0.3 (0.0)\\
M4-UR & 1.0 (\minus{}0.2) & \minus{}0.1 (0.4) & 0.2 (0.0) & 0.1 (0.1) & 0.9 (\minus{}0.2) & \minus{}0.5 (0.4) & 0.2 (0.1)\\
M4-Prompts & 3.6 (1.9) & 2.2 (0.7) & 1.8 (0.5) & 0.9 (0.2) & 2.4 (1.2) & 0.5 (\minus{}0.4) & 1.9 (0.6)\\
M4-Prompts FT & 3.2 (\minus{}0.1) & 1.9 (2.2) & 1.5 (\minus{}1.0) & 0.8 (1.4) & 2.0 (\minus{}0.7) & 0.5 (1.3) & 1.6 (0.5)\\
PaLM 8B & 6.5 (2.2) & 1.7 (1.0) & 4.6 (0.8) & 0.7 (0.4) & 4.3 (0.9) & 0.5 (0.1) & 2.8 (0.9)\\
PaLM 62B & 13.1 (4.0) & 5.2 (2.7) & 9.6 (1.7) & 2.2 (1.1) & 8.0 (1.2) & 2.7 (0.9) & 6.5 (1.9)\\
PaLM 540B & 13.8 (4.1) & 7.0 (3.2) & 9.7 (1.7) & 4.0 (1.5) & 9.1 (1.4) & 3.9 (1.5) & 7.7 (2.2)\\
\midrule
& zh-CN & zh-TW & zh-CN & zh-TW & zh-CN & zh-TW & zh\\\midrule
UR & 1.0 (\minus{}0.1) & \minus{}0.4 (0.2) & 1.0 (0.5) & \minus{}0.3 (\minus{}0.4) & 1.8 (1.1) & \minus{}0.9 (\minus{}0.8) & 0.2 (0.1)\\
M4-UR & \minus{}0.1 (\minus{}0.3) & 0.2 (0.3) & 0.3 (0.1) & 0.0 (\minus{}0.1) & \minus{}0.1 (\minus{}0.1) & \minus{}0.1 (0.2) & 0.0 (0.0)\\
M4-Prompts & 0.6 (1.6) & \minus{}0.5 (\minus{}1.8) & 1.3 (1.2) & \minus{}0.5 (\minus{}1.2) & 1.3 (2.0) & \minus{}0.7 (\minus{}1.4) & 0.1 (0.0)\\
M4-Prompts FT & 1.5 (1.0) & \minus{}0.7 (\minus{}0.9) & 2.0 (0.8) & \minus{}1.2 (\minus{}0.8) & 1.6 (1.0) & \minus{}1.2 (\minus{}1.1) & 0.1 (0.0)\\
PaLM 8B & 2.0 (1.1) & 1.9 (0.7) & 3.0 (1.0) & 0.2 (\minus{}0.9) & 2.4 (0.4) & 1.1 (0.2) & 1.7 (0.4)\\
PaLM 62B & 5.9 (1.7) & 3.5 (1.5) & 4.3 (0.8) & 1.2 (0.0) & 5.9 (1.1) & 0.2 (0.6) & 3.3 (0.9)\\
PaLM 540B & 9.8 (4.2) & 4.7 (1.8) & 6.4 (1.4) & 0.5 (0.0) & 9.0 (2.0) & \minus{}0.5 (0.4) & 4.7 (1.6)\\
\bottomrule
\end{tabular}
\caption{FRMT test set deltas between matched and mismatched outputs for a given reference, shown in the format: $\Delta$BLEU~($\Delta$BLEURT)\@. Negative numbers indicate that the reference-based metric preferred the model output that targeted the opposite language variety. The last column shows deltas between FRMT scores evaluated with respect to matched vs.~mismatched outputs.}
\label{tab:mismatched_outputs}
\end{table*}

We observe that in the Portuguese case, most models do score better when asked to produce text in the same regional variety as the reference. However, when it comes to Mandarin, most models---PaLM being the exception---struggle to produce zh-TW output that outperforms their zh-CN output when evaluated against a zh-TW reference, indicating that the attempts to appropriately stylize the generated text degrade its quality more than they improve its regional acceptability.

\subsection{Effect of exemplars} \label{sec:exemplar_ablations}

\begin{table*}
\centering
\footnotesize
\begin{tabular}{rcccccc|c}
\toprule
& \multicolumn{2}{c}{\bf Lexical} & \multicolumn{2}{c}{\bf Entity} & \multicolumn{2}{c}{\bf Random} & \bf FRMT \\
\bf Exemplars & pt-BR & pt-PT & pt-BR & pt-PT & pt-BR & pt-PT & pt\\\midrule

0 & 50.7 (75.7) & 35.6 (71.2) & 56.4 (80.3) & 47.4 (77.6) & 53.0 (76.0) & 42.4 (73.6) & 47.2 (75.7)\\
1 & 52.0 (\textbf{77.1}) & 39.7 (73.7) & 57.0 (81.2) & 49.1 (78.5) & 54.5 (\textbf{77.0}) & 45.1 (75.2) & 49.3 (77.1)\\
5 & 53.2 (77.0) & 40.0 (\textbf{74.0}) & 58.5 (81.2) & 48.6 (78.7) & 54.8 (76.8) & 45.2 (75.3) & 49.8 (77.2)\\
7 & 53.5 (\textbf{77.1}) & 40.0 (73.8) & 58.6 (\textbf{81.3}) & 48.8 (78.8) & \textbf{55.2} (\textbf{77.0}) & \textbf{45.8} (\textbf{75.5}) & 50.0 (77.2)\\
10 & \textbf{53.7} (\textbf{77.1}) & \textbf{40.1} (73.9) & \textbf{59.0} (81.2) & \textbf{49.5} (\textbf{79.0}) & 54.8 (76.9) & 45.6 (\textbf{75.5}) & \textbf{50.2} (\textbf{77.3})\\\midrule
& zh-CN & zh-TW & zh-CN & zh-TW & zh-CN & zh-TW & zh\\\midrule
0 & 32.7 (64.5) & 22.2 (61.3) & 40.3 (72.7) & 32.8 (70.2) & 38.7 (65.6) & 29.0 (63.1) & 32.3 (66.2)\\
1 & 35.1 (66.4) & 24.6 (\textbf{64.3}) & 43.7 (74.2) & 35.2 (\textbf{72.8}) & 39.9 (67.6) & 31.1 (66.4) & 34.6 (68.6)\\
5 & 35.1 (\textbf{66.7}) & 25.0 (63.9) & 44.7 (74.6) & \textbf{35.3} (\textbf{72.8}) & 40.0 (67.6) & \textbf{31.8} (\textbf{66.7}) & \textbf{35.0} (68.7)\\
7 & \textbf{35.4} (66.6) & \textbf{25.3} (64.2) & \textbf{45.3} (\textbf{74.7}) & 34.9 (72.6) & \textbf{40.7} (\textbf{68.0}) & 30.5 (66.6) & \textbf{35.0} (\textbf{68.8})\\
10 & 34.8 (66.5) & 24.6 (63.4) & 44.8 (\textbf{74.7}) & 35.2 (72.5) & 40.0 (67.8) & 29.6 (66.0) & 34.5 (68.4)\\
\bottomrule
\end{tabular}
\caption{FRMT test set results of PaLM 540B, when varying the number of exemplars, shown in the format: BLEU~(BLEURT)\@. Across both languages, even one exemplar is sufficient for strong results, and zero-shot performance is reasonably strong. Increasing to $10$ exemplars in Portuguese or $7$ exemplars in Mandarin gives marginal additional gains. Note that these results were not used to select the number of exemplars for the PaLM 540B results reported elsewhere; this ablation was run afterward.}
\label{tab:palm_exemplars}
\end{table*}

\begin{table*}
\centering
\footnotesize
\begin{tabular}{p{0.06\textwidth}p{0.42\textwidth}p{0.42\textwidth}}
\toprule
\textbf{Model} & \textbf{Target:pt-BR} & \textbf{Target:pt-PT} \\
\midrule
Gold & A legalização do casamento entre pessoas do mesmo sexo em Portugal ocorreu \textbf{em} 17 de maio de 2010. & O casamento entre pessoas do mesmo sexo foi legalizado em Portugal \textbf{a} 17 de maio de 2010. \\
\midrule
PaLM & O casamento entre pessoas do mesmo sexo em Portugal foi legalizado \textbf{em} 17 de maio de 2010. & O casamento entre pessoas do mesmo sexo em Portugal foi legalizado \textbf{a} 17 de Maio de 2010. \\
\midrule
M4-Prompts & O casamento entre pessoas do mesmo sexo em Portugal foi legalizado \textbf{em} 17 de maio de 2010. & O casamento entre pessoas do mesmo sexo em Portugal foi legalizado \textbf{a} 17 de maio de 2010. \\
\midrule
M4-UR & O casamento homoafetivo em Portugal foi legalizado \textbf{em} 17 de Maio de 2010. & O casamento homoafetivo em Portugal foi legalizado \textbf{a} 17 de Maio de 2010. \\
\bottomrule
\end{tabular}
\caption{Gold and model outputs for the source: \textit{Same-sex marriage in Portugal was legalized on 17 May 2010.} Phenomena of interest are \textbf{bolded}.}
\label{tab:examples_pt}
\end{table*}

\begin{table*}
\centering
\footnotesize
\resizebox{1.0\textwidth}{!}{
\begin{tabular}{p{0.1\textwidth}p{0.42\textwidth}p{0.42\textwidth}}
\toprule
\textbf{Model} & \textbf{Target:zh-CN} & \textbf{Target:zh-TW} \\
\midrule
Gold & \simp{并非所有的\textbf{软件}缺陷都是由代码错误导致的。} & \trad{並非所有\textbf{軟體}缺陷都是因\textbf{程式}碼錯誤所導致。} \\
\midrule
PaLM & \simp{并不是所有的\textbf{软件}缺陷都是由\textbf{编码}错误造成的。} & \trad{並不是所有的\textbf{軟體}缺陷都是由\textbf{程式}錯誤所造成。} \\
\midrule
M4-Prompts & \simp{并非所有的\textbf{软件}缺陷都是由\textbf{编码}错误引起的。} & \trad{並非所有的\textbf{\error{軟件}}缺陷是由\textbf{\error{編碼}}錯誤引起的。} 
\\
\midrule
M4-UR & \simp{并非所有的\textbf{软件}缺陷都是由\textbf{编码}错误引起的。} & \trad{並非所有的\textbf{\error{軟件}}缺陷都是由\textbf{\error{編碼}}錯誤引起的。} 
\\
\bottomrule
\end{tabular}
} 
\caption{Gold and model outputs for the source: \textit{Not all software defects are caused by coding errors.} Phenomena of interest are \textbf{bolded}, and region-specific errors are \textbf{\error{underlined and red}}. Note, M4-based model zh-TW outputs have been transliterated to traditional script, matching our evaluation setting.}
\label{tab:examples_zh}
\end{table*}

To test sensitivity to the number and choice of exemplars, we evaluate PaLM 540B while varying the set of exemplars used.  Table \ref{tab:palm_exemplars} shows the effect of ablating the number of exemplars in the range $0$--$10$. We observe that a single exemplar is sufficient to achieve strong results, using zero exemplars yields reasonably strong results, and gains from additional exemplars are marginal.

To measure the variance in performance across exemplar choice, we re-run PaLM 540B evaluation three times each using either $1$ or $10$ exemplars, resampling the exemplars on each run. We find that the choice of exemplar(s) has a relatively small effect---with $10$ exemplars, the standard deviations of FRMT-BLEU and FRMT-BLEURT across all four runs (including the original) were below $0.5$ in each language, and with just $1$ exemplar, the standard deviations remained under $1.0$.

\subsection{Qualitative analysis}\label{sec:results_qualitative}

To provide additional insights on regional differences and model behavior, we manually inspect dev set gold translations and model outputs, across the models sent to human evaluation. In both languages, we observe regional differences beyond just the lexical items underlying our \texttt{lexical} bucket. For instance, in Table~\ref{tab:examples_pt} and similar examples, we find \textit{on <date>} phrases tend to be translated with differing prepositions---\textit{em} in pt-BR and \textit{a} in pt-PT\@.  As another example, in Table~\ref{tab:examples_zh}, we observe both gold and PaLM outputs use the term \trad{程式} (chéngshì, en:\textit{program}) only in zh-TW when translating the phrase ``coding errors’’.

In many cases, PaLM uses the expected region-specific lexical forms, as already reflected in our lexical accuracy metric. By contrast, we observe the M4-based models are more prone to use terms from the web-majority region (pt-BR and zh-CN) irrespective of the target. For example, in Table~\ref{tab:examples_zh}, PaLM matches gold translations in using the region-specific terms for software---zh-CN: \simp{软件} (ruǎnjiàn), zh-TW: \trad{軟體} (ruǎntǐ)---while the M4-based models use the zh-CN term throughout (simplified: \simp{软件}, traditional: \trad{軟件}).

%% file: 7-conclusion.tex
\section{Conclusion}\label{sec:conclusion}

In this paper, we introduced FRMT, a new benchmark for evaluating few-shot region-aware machine translation.  Our dataset covers $4$ regions of Portuguese and Mandarin, and enables fine-grained comparison across region-matched and mismatched conditions, and across different classes of inputs (\texttt{lexical}, \texttt{entity}, \texttt{random}).

While we found the large-scale generalist model PaLM 540B to show impressive few-shot region control, there is still significant room for improvement.  None of the models we evaluated match human performance, and the gap is particularly large in Mandarin.  Additionally, there remains an open research question as to whether robust few-shot regional control can be achieved at more modest model scales.

We are eager to see progress on FRMT, as methods that do well in this few-shot setting are likely to be easily extensible to other regions and styles. We anticipate that the flexibility to adapt to new output styles in the absence of extensive labeled data will be a key factor in making generative text models more useful, inclusive, and equitable.